\documentclass[runningheads]{llncs}
\usepackage[T1]{fontenc}
\usepackage{graphicx}
\usepackage{graphicx}
\usepackage{textcomp}
\usepackage{multirow}

\usepackage{subfig}
\usepackage{flushend}
\usepackage{booktabs}
\usepackage{graphicx}
\usepackage{amsmath, amssymb, amsfonts}
\usepackage{float}
\usepackage[font=small]{caption}   

\usepackage[hidelinks]{hyperref}
\begin{document}
\title{DAStatFormer: A Hybrid Multibranch Transformer with Statistical Feature Integration for DAS-Based Pattern Recognitions}

\titlerunning{DAStatFormer for DAS-based Pattern Recognitions}
%
\author{Michel Dione\inst{1} \and
Jerry Lonlac\inst{1} \and
Hélène Louis\inst{1} \and
Anthony Fleury\inst{1} \and
Stéphane Lecoeuche\inst{2}}
\authorrunning{M. Dione et al.}
%
\institute{$^1$ IMT Nord Europe, Institut Mines-Telecom,
Univ. Lille, Centre for Digital Systems
Lille, France\\
$^2$ IMT Mines Ales, Institut Mines-Telecom, Ales, France\\
\email{michel.dione@imt-nord-europe.fr}} 

%
\maketitle              
%

\begin{abstract}
Distributed Acoustic Sensing (DAS) enables large-scale monitoring through optical fibers, but its high dimensionality and complex spatio-temporal patterns make event classification demanding. Existing deep learning approaches--CNNs, recurrent models, and Transformer variants--either fail to capture long-range dependencies or require processing raw DAS matrices at prohibitive cost.
We propose \textit{DAStatFormer}, a hybrid multibranch Transformer that combines compact multidomain statistical features with Gated Transformer Networks. Instead of raw signals, we extract 24 ANOVA-selected attributes per channel from the temporal, waveform, and spectral domains, reducing data size by orders of magnitude while preserving discriminative information. Each domain is processed via dedicated step-wise and channel-wise attention branches, fused by an adaptive gating mechanism.
Experiments on the open $\phi$-OTDR benchmark and a real-scenario DAS dataset show that DAStatFormer achieves up to 99.4\% accuracy and near-perfect real-world performance, while using significantly fewer parameters and lower inference cost than models such as DASFormer and DeepViT. These results demonstrate its suitability for scalable, real-time DAS-based monitoring.\\
We release our code at \url{https://github.com/MichelD-git/DAStatFormer}


\end{abstract}

\section{Introduction}

Distributed Acoustic Sensing (DAS) has become a powerful solution for large-scale monitoring of infrastructures, transportation networks, and sensitive perimeters~\cite{ghazali2024state}.
By converting optical fibers into dense arrays of virtual sensors, DAS enables real-time detection and localization of diverse activities, even in challenging environmental conditions.
Deep learning has markedly advanced DAS-based event recognition, with CNNs~\cite{jiang2018event,meng2020research}, hybrid CNN-BiLSTM models~\cite{wu2020novel}, and Transformer-based architectures such as DASFormer~\cite{li2025dasformer}.
However, these models still face key limitations: they struggle to simultaneously capture fine-grained temporal patterns and broader statistical trends, and processing raw spatio-temporal DAS matrices remains computationally demanding due to extremely high sampling rates leading to scalability issues in memory usage, training cost, and inference speed.
To overcome these challenges, we introduce \textit{DAStatFormer}, a hybrid Transformer-based framework that integrates multidomain statistical feature extraction with a multibranch Gated Transformer Network (GTN) architecture~\cite{liu2021gated}. Instead of operating on raw DAS matrices, the method computes 24 interpretable attributes per channel across three complementary domains temporal, waveform, and spectral following principles from acoustic and vibration analysis. These compact descriptors preserve discriminative information while drastically reducing dimensionality, and are processed by three domain-specific Transformer branches with two-way gating mechanisms that balance step-wise and channel-wise dependencies. This design enables efficient modeling of both local and global structures in DAS data at low computational cost.
The proposed framework is evaluated on two datasets: (1) the open $\phi$-OTDR dataset~\cite{cao2023open}, and (2) a real-scenario DAS dataset introduced by Tomasov \textit{et al.}~\cite{tomasov2025comprehensive}, which we restructure into a windowed format compatible with learning-based models. Experiments show that \textit{DAStatFormer} achieves accuracy comparable to state-of-the-art approaches while requiring significantly fewer parameters and reduced inference time. The main contributions of this work are:

\begin{enumerate}
    \item A hybrid Transformer architecture integrating multidomain statistical features with a multibranch GTN-based attention mechanism for DAS event recognition.
    \item A compact and interpretable 24-feature representation extracted from temporal, waveform, and spectral domains, offering strong discriminability with minimal data dimensionality.
    \item A unified preprocessing pipeline restructuring real DAS recordings into a standardized windowed format suitable for Transformer models.
    \item Extensive validation on laboratory and real-world datasets, demonstrating strong accuracy, efficiency, and scalability for real-time DAS monitoring.
\end{enumerate}
The remainder of this paper is organized as follows: Section~\ref{sec:related work} reviews prior work on DAS event recognition, and Section~\ref{sec:das} summarizes the principles of DAS. Section~\ref{sec:methodology} presents the proposed DAStatFormer framework, Section~\ref{sec:results} reports experimental evaluations, and Section~\ref{sec:conclusion} concludes the study.

\section{Related Work}\label{sec:related work}

Distributed Acoustic Sensing (DAS) has become a key technology for applications such as pipeline monitoring, geotechnical analysis, and infrastructure surveillance~\cite{ghazali2024state}. Its growing use for perimeter intrusion detection has stimulated rapid progress in signal processing and machine learning. Early methods based on handcrafted features and classical classifiers provided useful insights but proved insufficient to handle the high dimensionality, noise levels, and strong spatio-temporal correlations that characterize DAS signals.
To overcome these limitations, researchers explored both traditional machine learning models-such as Support Vector Machines (SVM)~\cite{jia2019knn}, Random Forests, and gradient-boosted trees and more advanced deep learning architectures. Convolutional Neural Networks (CNNs) and Recurrent Neural Networks (RNNs), including LSTM variants, have become widely used thanks to their ability to learn discriminative features directly from raw or preprocessed DAS signals~\cite{tejedor2017machine}.  
For instance, Cao \textit{et al.}~\cite{cao2023open} employed a 2D CNN to classify events in a laboratory DAS setup, successfully identifying activities such as digging and walking.
Similarly, Wu \textit{et al.}~\cite{wu2020novel} proposed a 1D CNN-BiLSTM hybrid model that jointly captures spatial and temporal dependencies, improving classification accuracy in intrusion scenarios.
Other studies reformulated DAS event recognition as a visual classification task by converting spatio-temporal signals into waterfall diagrams~\cite{bublin2021event} or spectrograms~\cite{guzhov2021esresnet}, enabling the use of computer vision models.
However, this transformation often discards subtle signal details, making such approaches suitable only when the temporal and spatial characteristics of the data are relatively uniform.
In the context of railway intrusion monitoring with $\phi$-OTDR, Meng \textit{et al.}~\cite{meng2020research} introduced an XGBoost-based identification method, showing the potential of gradient-boosted trees for DAS event recognition.
Overall, CNN-based models have become a dominant strategy for processing $\phi$-OTDR data, with 1D and 2D variants widely adopted.
Despite their success, CNN architectures present intrinsic limitations for DAS analysis: they focus mainly on local patterns, require fixed-size inputs, and struggle to capture global contextual information~\cite{liu2021gated}.
Because DAS recordings are extremely high-dimensional and contain long-range spatio-temporal dependencies along the fiber, CNNs are not ideally suited to fully model such complex structures, which ultimately restricts their effectiveness.
These constraints have motivated the investigation of alternative architectures capable of modeling long-range interactions.\\
Transformer-based models, originally introduced for natural language processing~\cite{vaswani2017attention}, have recently been adapted to time-series and sensing tasks thanks to their powerful self-attention mechanism. In the DAS domain, this has inspired works such as DASFormer~\cite{li2025dasformer}, which directly model spatio-temporal dependencies; and DeepViT~\cite{zhou2021deepvit}, building on the success of Vision Transformers (ViT)~\cite{song2025facial} by applying them to waterfall-diagram representations for intrusion detection. Dione \textit{et al.}~\cite{dione2025intrusion} introduced \textit{DASViT1D}, a Transformer-based model leveraging time--frequency representations for DAS intrusion recognition.
However, existing Transformer-based DAS approaches still face notable challenges: most operate on raw high-dimensional spatio-temporal matrices or on visually transformed images, both of which demand substantial computational resources and may compromise signal fidelity. 
Furthermore, few studies explicitly exploit the multidomain nature of DAS signals combining temporal, waveform, and spectral information within a unified learning framework.  
Addressing these gaps, our work introduces \textit{DAStatFormer}, a hybrid multibranch Transformer architecture that integrates domain-specific statistical features with an efficient attention-based modeling strategy.  
This design achieves a balance between dimensionality reduction, computational scalability, and high discriminative performance, paving the way for more interpretable and resource-efficient DAS-based monitoring systems.

\section{Distributed Acoustic Sensing (DAS) System} \label{sec:das}
\begin{figure}
\centering
    \includegraphics[width=6.5cm]{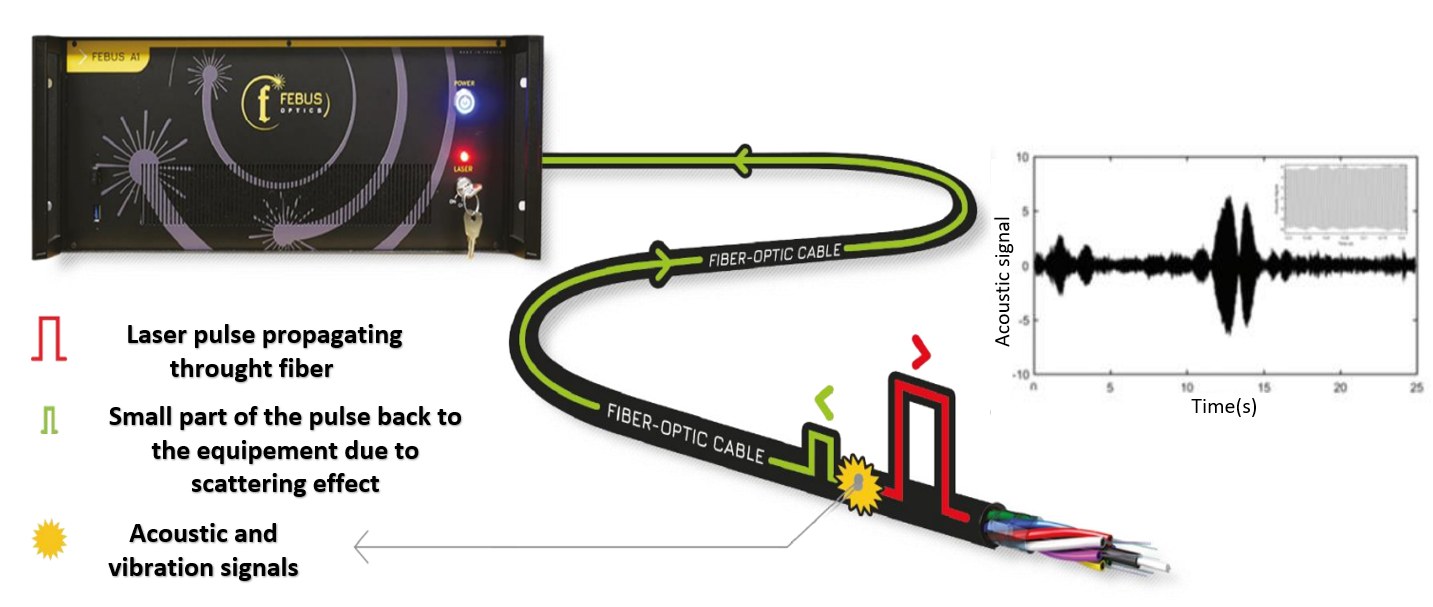}
    \caption{Distributed Acoustic Sensing Principle \cite{jestin2020integration}}
\label{das_principle}
\end{figure}

Distributed Acoustic Sensing (DAS) exploits Rayleigh backscattering to detect vibrations along optical fibers over long distances~\cite{xie2023label}.  
Short optical pulses are injected into the fiber, and external acoustic disturbances induce small strain variations that modulate the phase and intensity of the backscattered light.  
These variations are sampled at discrete locations, referred to as \textit{channels} or virtual sensors, whose spacing is determined by the system spatial resolution~\cite{jiang2018event}.  
The DAS interrogator continuously records these measurements, producing dense spatio-temporal strain signals along the monitored fiber.\\
A DAS system is mainly characterized by its sampling frequency $f_s$, spatial resolution $dx$, fiber length $fl$, and number of channels $n$.  
Figure~\ref{das_principle} illustrates the general operating principle of a DAS system.
\vspace{-0.4cm}
\vspace{-0.6cm}
A DAS acquisition produces a high-dimensional spatio-temporal signal that can be naturally represented as a matrix of strain measurements indexed by time and space.  
Each DAS recording consists of $m = f_s \times T$ temporal samples acquired along $n$ spatial channels distributed along the optical fiber, where $f_s$ denotes the sampling frequency and $T$ the recording duration.  
At each time instant $t_i$, the DAS system measures a strain value $str_i^j$ at channel $\mathrm{Ch}_j$, corresponding to a fixed spatial position along the fiber.  
This organization results in a dense spatio-temporal representation in which rows correspond to time instants and columns to spatial sensing channels.  
Such high-resolution DAS data capture the dynamic response of the monitored environment along the entire fiber and form the basis for subsequent feature extraction, signal representation, and event classification.







\section{Methodology} \label{sec:methodology}

DAS recordings are extremely high-dimensional due to the joint effects of high sampling frequency, long acquisition duration, and dense spatial resolution.
This results in large spatio-temporal matrices that are costly to process and difficult to scale.
The proposed \textit{DAStatFormer} framework addresses this challenge through a two-stage strategy.
First, DAS segments are transformed into compact multidomain statistical descriptors that capture the essential temporal, waveform, and spectral characteristics while substantially reducing data volume.
Second, the resulting features are processed by a multibranch Transformer architecture specifically designed to exploit cross-domain complementarity and to model both local and global dependencies in DAS signals despite the reduced dimensionality.\\
The following subsections describe the feature extraction process and the \textit{DAStatFormer} architecture.

\subsection{Feature Extraction}

Feature extraction aims to convert raw high-dimensional DAS traces into compact, informative representations suitable for learning-based analysis.
Rather than processing the full spatio-temporal matrices, we compute 24 statistical attributes distributed across three complementary domains: \textit{time}, \textit{waveform}, and \textit{spectral}.
Time-domain descriptors characterize the global amplitude distribution and signal dynamics; waveform-based features capture shape-related and impulsive behaviors; and spectral attributes describe frequency-dependent patterns crucial for distinguishing continuous activities from short transient events. The selected 24 attributes originate from a larger pool of \(60\) candidate features as proposed by C. Hibert et al.~\cite{hibert2017automatic}. To ensure objective relevance, we applied an ANOVA-based feature selection, retaining only the descriptors exhibiting high inter-class variance and low intra-class variability across all event categories.
This statistically grounded selection guarantees that each feature contributes meaningful discriminative information.
By combining temporal, morphological, and spectral cues, the resulting multidomain representation achieves a strong balance between expressiveness and dimensionality reduction.\\
This compact feature space preserves the essential properties of DAS signals while enabling efficient downstream processing within the proposed Transformer-based framework~\cite{chen2020review}.

Let the raw input signal be denoted as $x = \{x_1, x_2, \ldots, x_n\}$ of length $n$. The complete list of extracted attributes is provided in Table~\ref{tab:features}.

\begin{table}[!h]
\centering
\caption{Selected statistical attributes extracted from DAS signals.}
\label{tab:features}
\renewcommand{\arraystretch}{1.5}
\scriptsize
\begin{tabular}{|c|c|p{8cm}|}
\hline
\textbf{Domain} & \textbf{Attribute} & \textbf{Definition} \\
\hline
\multirow{7}{*}{Time-domain} 
 & $x_{\max}$ & Maximum amplitude of the signal \\
 & Peak-to-peak & $x_{\max} - x_{\min}$ \\
 & Std. deviation & $\sigma = \sqrt{\tfrac{1}{n}\sum (x_i-\mu)^2}$ \\
 & Skewness & $\tfrac{1}{n}\sum \left(\tfrac{x_i-\mu}{\sigma}\right)^3$ \\
 & Kurtosis & $\tfrac{1}{n}\sum \left(\tfrac{x_i-\mu}{\sigma}\right)^4$ \\
 & Zero-crossing rate & Mean rate of sign changes in $x(t)$ \\
 & Envelope ratio & $\displaystyle \frac{\max(e(t))}{\operatorname{mean}(e(t))}$, with $e(t)=|\mathcal{H}\{x(t)\}|$ \\
\hline
\multirow{6}{*}{Waveform-domain} 
 & Waveform factor & $\displaystyle \frac{\text{RMS}}{\operatorname{mean}(|x_i|)}$ \\
 & Impulse factor & $\displaystyle \frac{x_{\max}}{\operatorname{mean}(|x_i|)}$ \\
 & Crest factor & $\displaystyle \frac{x_{\max}}{\text{RMS}}$ \\
 & Kurtosis factor & $\displaystyle \frac{\tfrac{1}{n}\sum x_i^4}{\text{RMS}^4}$ \\
 & Envelope skewness & Skewness of the Hilbert envelope $e(t)$ \\
 & Envelope kurtosis & Kurtosis of the Hilbert envelope $e(t)$ \\
\hline
\multirow{11}{*}{Spectral-domain} 
 & Mean spectrum & $\displaystyle \mu_f = \tfrac{1}{M}\sum |X(f_k)|$ \\
 & Max spectrum & $\max_k |X(f_k)|$ \\
 & Spectral entropy & $H = - \sum p_k \log_2(p_k+\epsilon)$, $p_k=\tfrac{|X(f_k)|}{\sum_j |X(f_j)|}$ \\
 & Spectral centroid & $\displaystyle C = \frac{\sum_k f_k |X(f_k)|}{\sum_k |X(f_k)|}$ \\
 & Spectral bandwidth & $\displaystyle BW = \sqrt{\frac{\sum_k (f_k-C)^2 |X(f_k)|}{\sum_k |X(f_k)|}}$ \\
 & Spectral skewness & Third standardized moment of $|X(f_k)|$ around $C$ \\
 & Spectral kurtosis & Fourth standardized moment of $|X(f_k)|$ around $C$ \\
 & Spectral flatness & $\exp\big(\operatorname{mean}(\log |X(f_k)|)\big)\big/ \operatorname{mean}(|X(f_k)|)$ \\
 & Spectral slope & Linear regression slope of $|X(f_k)|$ as a function of $f_k$ \\
 & Rolloff 85\% & Frequency $f_{0.85}$ such that $\sum_{f_k \le f_{0.85}} |X(f_k)| = 0.85 \sum_k |X(f_k)|$ \\
 & Rolloff 95\% & Frequency $f_{0.95}$ such that $\sum_{f_k \le f_{0.95}} |X(f_k)| = 0.95 \sum_k |X(f_k)|$ \\
\hline
\end{tabular}
\end{table}


The waveform- and spectrum-based attributes are computed from the raw DAS segment $x(t)$, its Hilbert envelope $e(t)$, and its discrete Fourier transform (DFT) $X(f)$. The main notations are:
\begin{itemize}
    \item $e(t) = |\mathcal{H}\{x(t)\}|$: Hilbert envelope of the signal.
    \item $X(f) = \mathcal{F}\{x(t)\}$: discrete Fourier transform.
    \item $\mu$, $\sigma^2$, $\sigma$: mean, variance, and standard deviation.
    \item Skewness: $\tfrac{1}{n}\sum \big(\tfrac{x_i-\mu}{\sigma}\big)^3$.
    \item Kurtosis: $\tfrac{1}{n}\sum \big(\tfrac{x_i-\mu}{\sigma}\big)^4$.
    \item RMS: $\sqrt{\mu^2 + \sigma^2}$.
    \item Crest factor: $\tfrac{\max(x)}{\mathrm{RMS}}$.
    \item Spectral entropy: $H=-\sum p_i\log_2(p_i+\epsilon)$, with 
    $p_i=\tfrac{|X(f_i)|}{\sum_j |X(f_j)|}$.
\end{itemize}

A key motivation for using statistical descriptors is the drastic reduction of the temporal dimensionality of DAS recordings without sacrificing event discriminability.  
High DAS sampling rates generate extremely large spatio-temporal matrices, even for short signals. For example, a raw segment of size $10000 \times 12$ can be compactly reduced to a $12 \times 24$ representation by replacing each channel with 24 informative attributes, leading to substantial savings in computation and memory.
Each statistic is computed independently per DAS channel, yielding ordered spatial feature sequences along the fiber.  
Although spatial in nature, these sequences can be modeled analogously to temporal signals, making them well suited for Transformer-based architectures.  
Based on this representation, the next subsection introduces \textit{DAStatFormer}, a multibranch Transformer designed for efficient and accurate DAS event recognition.

\subsection{DAStatFormer model}

As described in the previous subsection, the first stage of our framework converts raw distributed acoustic sensing (DAS) traces into multidomain statistical descriptors spanning the temporal, waveform, and spectral domains. These compact yet expressive representations serve as inputs to the attention-based model presented below. Figure~\ref{fig:arch_transformer4d} provides an overview of the proposed architecture: three parallel branches process feature matrices extracted from the time, waveform, and spectral domains, and their outputs are adaptively fused for final classification.

\begin{figure}[t]
\centering
\includegraphics[width=9cm]{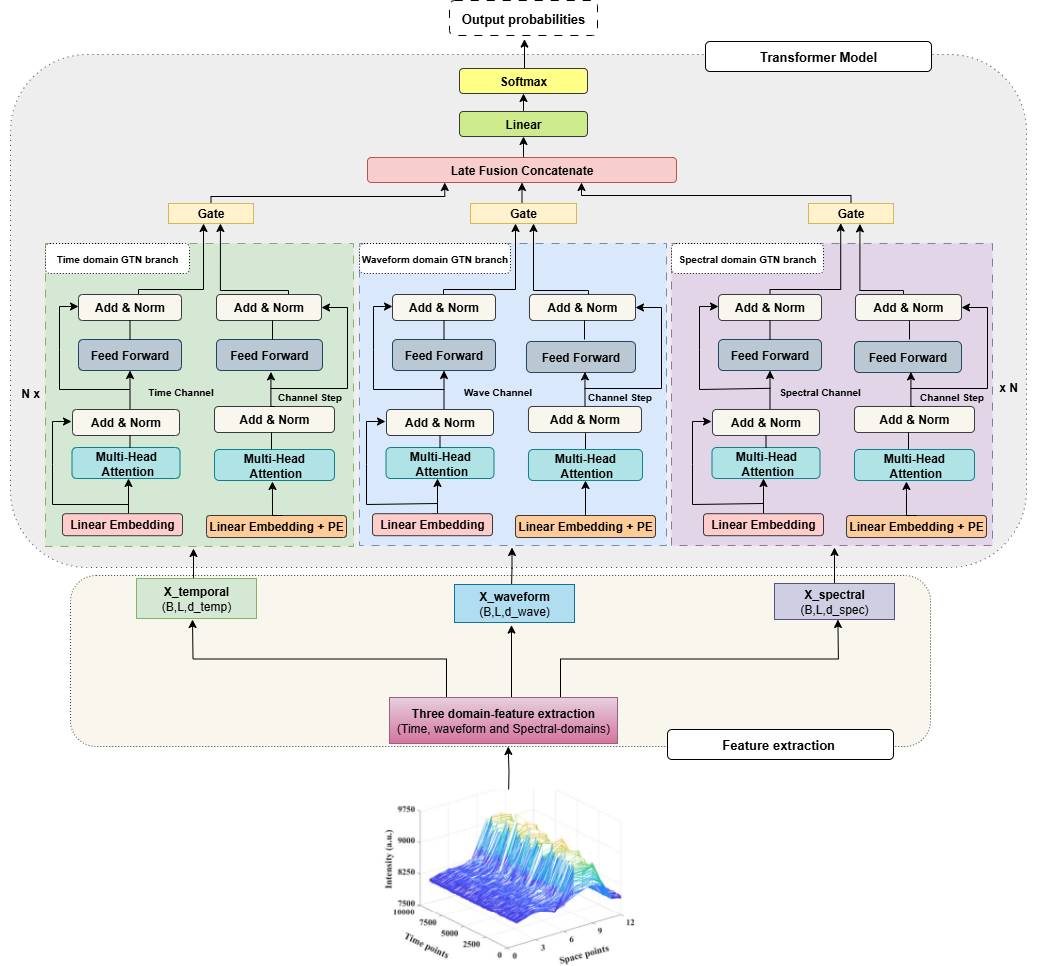}
\caption{Overview of the proposed \textit{DAStatFormer} framework.}
\label{fig:arch_transformer4d}
\end{figure}
\vspace{-0.23cm}
\paragraph{\textbf{Input representation}}

Let \(d \in \{\mathrm{time},\,\mathrm{waveform},\,\mathrm{spectral}\}\) denote one of the three feature domains.  
The multidomain descriptors for domain \(d\) are represented as a third-order tensor \(X^{(d)} \in \mathbb{R}^{B \times L \times d_{\mathrm{attr}}^{(d)}}\), where \(B\) is the mini-batch size, \(L\) is the number of sensing channels along the DAS fibre (interpreted as the sequence length), and \(d_{\mathrm{attr}}^{(d)}\) is the number of statistical attributes extracted per channel.  
We denote by \(x_{b,l}^{(d)} \in \mathbb{R}^{d_{\mathrm{attr}}^{(d)}}\) the feature vector corresponding to batch index \(b\) and channel \(l\) in domain \(d\).
\vspace{-0.4cm}
\subsubsection{Step-wise encoding}
The first path within each branch models spatial (inter-channel) correlations along the sensing fibre.  
Each attribute vector \(x_{b,l}^{(d)}\) is projected into a \(d_{\mathrm{model}}\)-dimensional embedding through a learnable linear transformation:
\begin{equation}
\mathbf{s}_{b,l}^{(d)} = W_{\mathrm{step}}^{(d)}\, x_{b,l}^{(d)} + b_{\mathrm{step}}^{(d)},
\qquad W_{\mathrm{step}}^{(d)} \in \mathbb{R}^{d_{\mathrm{attr}}^{(d)}\times d_{\mathrm{model}}},
\end{equation}
where \(W_{\mathrm{step}}^{(d)}\) and \(b_{\mathrm{step}}^{(d)}\) are trainable parameters.  
An optional sinusoidal positional encoding \(\mathrm{PE}\in\mathbb{R}^{L\times d_{\mathrm{model}}}\) is added to preserve the spatial ordering of channels.  
The resulting sequence \(S^{(d)}=[\mathbf{s}_{b,1}^{(d)},\ldots,\mathbf{s}_{b,L}^{(d)}]\in\mathbb{R}^{B\times L\times d_{\mathrm{model}}}\) is processed by \(N\) stacked Transformer encoder layers, each composed of multi-head self-attention (MHA) and a position-wise feed-forward network (FFN) with residual connections and layer normalization.
\vspace{-0.1cm}
\begin{itemize}
\item \textbf{Multi-head self-attention}  
Given queries \(Q\), keys \(K\), and values \(V\) (all equal to \(S^{(d)}\)), the \(i\)-th attention head computes:
\begin{equation}
\mathbf{h}_i = f\bigl(W_i^{(q)} Q,\, W_i^{(k)} K,\, W_i^{(v)} V\bigr),
\end{equation}
where \(W_i^{(q)}, W_i^{(k)}, W_i^{(v)} \in \mathbb{R}^{(d_{\mathrm{model}}/h)\times d_{\mathrm{model}}}\) are learnable projection matrices, \(h\) is the number of heads, and \(f\) denotes scaled dot-product attention.  
The outputs from all heads are concatenated and linearly projected:
\begin{equation}
\mathrm{MHA}(S^{(d)}) = W^{(o)}[\mathbf{h}_1;\ldots;\mathbf{h}_h],
\qquad W^{(o)} \in \mathbb{R}^{d_{\mathrm{model}}\times (h\,d_{\mathrm{model}}/h)}.
\end{equation}

\vspace{-0.2cm}
\item \textbf{Position-wise feed-forward network}  
Each position vector \(\mathbf{s} \in \mathbb{R}^{d_{\mathrm{model}}}\) is independently transformed by an FFN:
\begin{equation}
\mathrm{FFN}(\mathbf{s}) = \sigma(\mathbf{s} W_1 + b_1)\, W_2 + b_2,
\end{equation}
where \(W_1 \in \mathbb{R}^{d_{\mathrm{model}}\times d_{\mathrm{hidden}}}\),
\(W_2 \in \mathbb{R}^{d_{\mathrm{hidden}}\times d_{\mathrm{model}}}\),
and \(\sigma\) is a non-linear activation (ReLU in our implementation).  
Residual connections and layer normalization follow both the attention and FFN sublayers.  
After \(N\) encoder layers, the sequence \(S^{(d)}\) is flattened along the channel dimension to yield \(\mathrm{s\_flat}^{(d)} \in \mathbb{R}^{B\times (L\,d_{\mathrm{model}})}\).
\end{itemize}

\vspace{-0.55cm}
\subsubsection{Channel-wise encoding}

To capture dependencies among statistical attributes, a second path transposes the input tensor so that attributes form the sequence dimension: \(\tilde{X}^{(d)} \in \mathbb{R}^{B\times d_{\mathrm{attr}}^{(d)}\times L}\).  
Each slice \(\tilde{x}_{b,j}^{(d)}\in\mathbb{R}^{L}\) (with \(j\) indexing attributes) is embedded via:
\begin{equation}
\mathbf{c}_{b,j}^{(d)} = W_{\mathrm{chan}}^{(d)}\, \tilde{x}_{b,j}^{(d)} + b_{\mathrm{chan}}^{(d)},
\qquad W_{\mathrm{chan}}^{(d)} \in \mathbb{R}^{L\times d_{\mathrm{model}}}.
\end{equation}
The resulting sequence \(C^{(d)}\in\mathbb{R}^{B\times d_{\mathrm{attr}}^{(d)}\times d_{\mathrm{model}}}\) is passed through the same stack of \(N\) Transformer encoders to capture inter-attribute dependencies.  
It is then flattened to \(\mathrm{c\_flat}^{(d)} \in \mathbb{R}^{B\times (d_{\mathrm{attr}}^{(d)}\,d_{\mathrm{model}})}\).
\vspace{-0.4cm}
\subsubsection{Intra-branch gating and fusion}

To adaptively balance temporal and attribute-wise information, each branch employs a soft gating mechanism inspired by Gated Transformer Networks.  
The flattened vectors \(\mathrm{s\_flat}^{(d)}\) and \(\mathrm{c\_flat}^{(d)}\) are concatenated and fed to a two-unit softmax gate:
\begin{equation}
\mathbf{g}^{(d)} = \operatorname{softmax}\!\Big(W_{\mathrm{gate}}^{(d)} [\mathrm{s\_flat}^{(d)};\mathrm{c\_flat}^{(d)}] + b_{\mathrm{gate}}^{(d)}\Bigr),
\end{equation}
where \(W_{\mathrm{gate}}^{(d)}\in\mathbb{R}^{2\times (L\,d_{\mathrm{model}} + d_{\mathrm{attr}}^{(d)}\,d_{\mathrm{model}})}\).  
The fused representation for domain \(d\):
\begin{equation}
\mathrm{fused}^{(d)} = [\, g^{(d)}_1 \cdot \mathrm{s\_flat}^{(d)} \;;\; g^{(d)}_2 \cdot \mathrm{c\_flat}^{(d)} \,].
\end{equation}
\vspace{-1cm}
\subsubsection{Late fusion and classification}

The outputs of the three domain-specific branches are concatenated to form a global representation:
\begin{equation}
\mathrm{fused}_{\mathrm{all}} = [\, \mathrm{fused}^{(\mathrm{time})} ; \mathrm{fused}^{(\mathrm{waveform})} ; \mathrm{fused}^{(\mathrm{spectral})} \,] 
\in \mathbb{R}^{B\times D_{\mathrm{fuse}}},
\end{equation}
where \(D_{\mathrm{fuse}}\) is the total dimensionality of the concatenated vectors.  
A final linear projection followed by a softmax layer produces the posterior probabilities over event classes:
\begin{equation}
\mathbf{y} = \operatorname{softmax}\!\big(W_{\mathrm{out}}\, \mathrm{fused}_{\mathrm{all}} + b_{\mathrm{out}}\big),
\qquad W_{\mathrm{out}}\in\mathbb{R}^{D_{\mathrm{fuse}}\times d_{\mathrm{out}}}.
\end{equation}

The proposed \textit{DAStatFormer} architecture combines self-attention ability to model long-range dependencies with interpretable, domain-specific statistical features.  
The step-wise path captures inter-channel (spatial-temporal) dynamics, while the channel-wise path encodes correlations among statistical attributes.  
The gating mechanism adaptively re-weights these complementary representations, and the final multidomain fusion leverages cross-domain complementarity between temporal, waveform, and spectral cues-yielding a compact, discriminative, and physically interpretable representation for DAS event classification.

\section{Experiments and Results} \label{sec:results}
\subsection{Baseline datasets}
To evaluate the performance and generalization capability of the proposed \textit{DAStatFormer} framework, experiments were conducted on two different DAS datasets covering both laboratory and real-world scenarios. Each dataset contains multiple event types representative of typical perimeter monitoring conditions. All recordings were acquired using a phase-sensitive optical time-domain reflectometer ($\phi$-OTDR) system. The diversity in data acquisition setups, noise conditions, and event types ensures a robust validation of the proposed approach. 
\vspace{-0.4cm}
\subsubsection{Laboratory $\phi$-OTDR Dataset}
\begin{figure} [!h]
\centering
\begin{minipage} {0.48\textwidth}
    \centering
    \includegraphics[width=1.\linewidth]{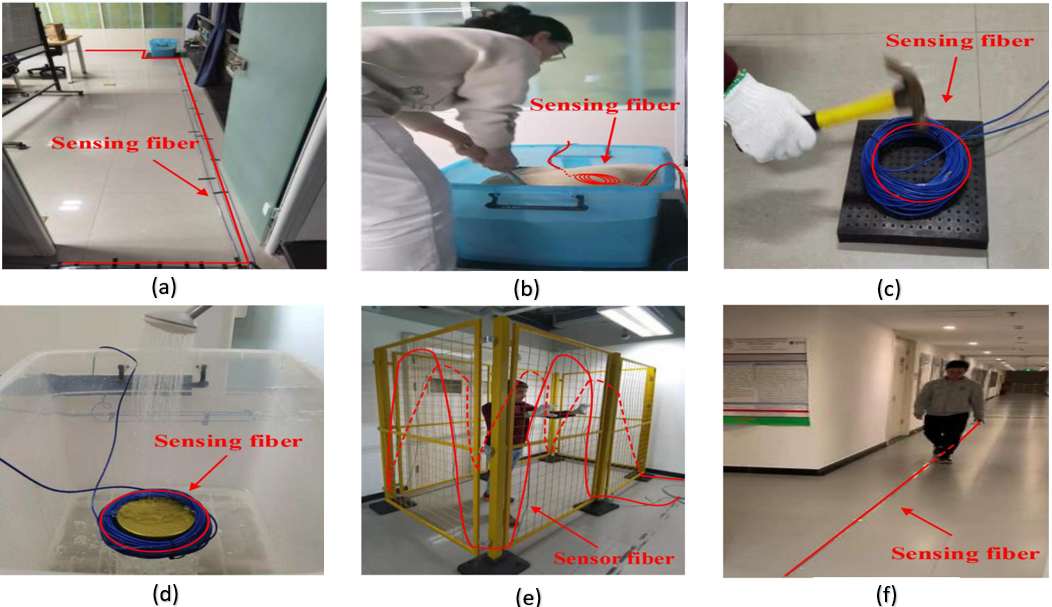}
    \caption*{\scriptsize (1) }
\label{fig:labo_scenario}
\end{minipage}
\hfill
\begin{minipage} {0.45\textwidth}
    \centering
    \includegraphics[width=\linewidth]{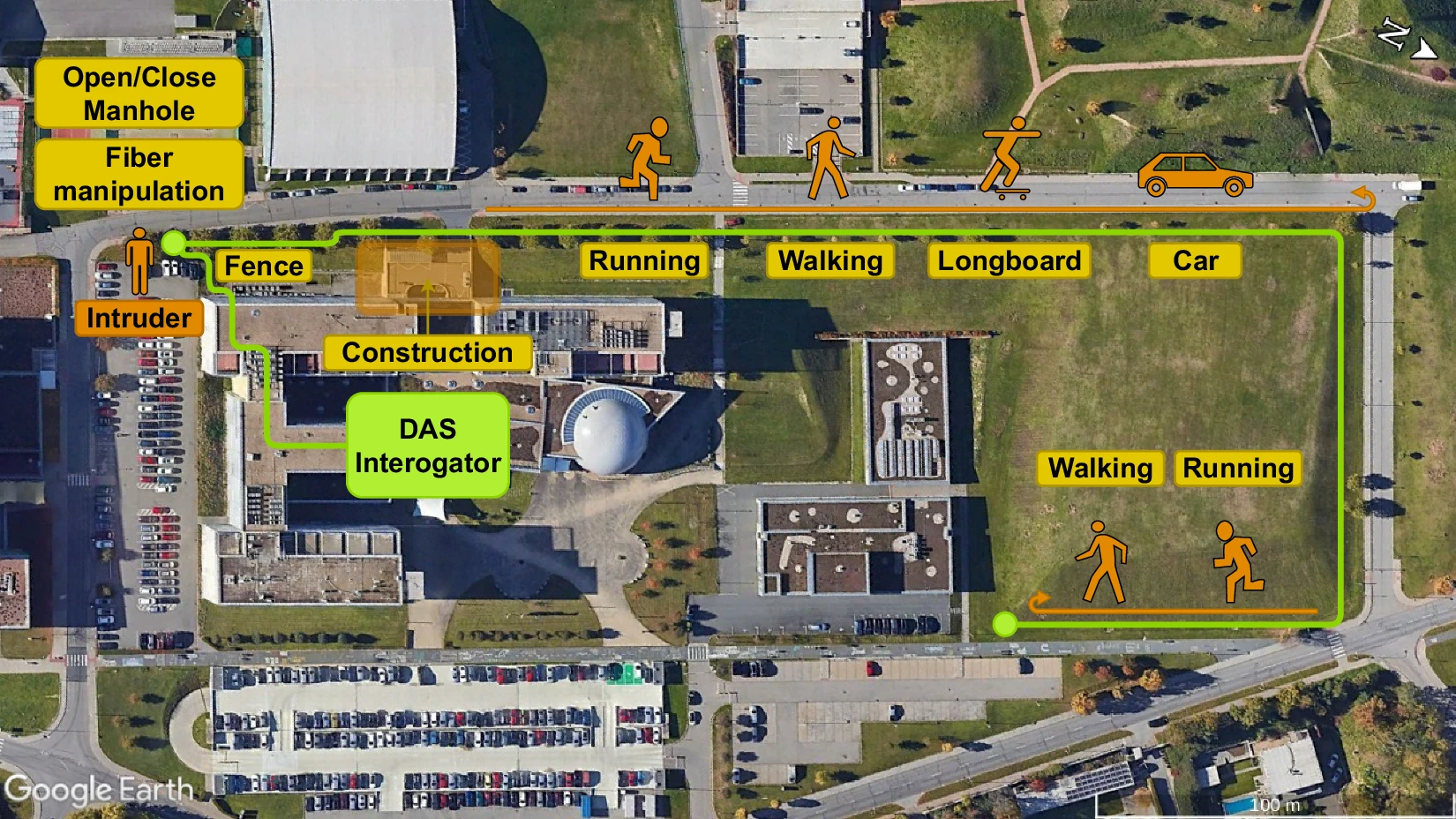}
    \caption*{\scriptsize (2) }
    \label{fig:univ_scenario}
\end{minipage}
\caption{ Overview of the two experimental configurations: 
\textbf{(1)} Laboratory $\phi$-OTDR setup illustrating six representative events (background, digging, knocking, watering, shaking, walking), with the sensing fiber highlighted in red~\cite{cao2023open}.  
\textbf{(2)} Aerial view of the campus DAS deployment, where the buried fiber (green line) monitors activities such as walking, running, longboarding, driving, fence climbing, and fiber manipulation~\cite{tomasov2025comprehensive}.}

\label{fig:datasets_overview}
\end{figure}

This dataset originates from a controlled laboratory experiment provided and described by Cao et al. in~\cite{cao2023open}, where a $\phi$-OTDR system was employed to monitor a buried optical fiber subjected to various mechanical perturbations.  
The experimental setup simulates a perimeter intrusion monitoring scenario under repeatable and noise-controlled conditions, as showed in Fig ~\ref{fig:datasets_overview}-1.  
Six distinct types of events were generated: \textit{background}, \textit{digging}, \textit{knocking}, \textit{watering}, \textit{shaking}, and \textit{walking}.  
Each event was performed multiple times at different spatial positions along the fiber to ensure sufficient variability and robustness of the dataset.  
Table~\ref{tab:event_numbers} summarizes the number of samples per class, along with their train/test split.  

\vspace{0.1cm}

\begin{table} [!h]
\centering
\begin{minipage} {0.43\linewidth}
\centering
\caption{ $\phi$-OTDR system config~\cite{cao2023open}.}
\label{tab:das_config}
\renewcommand{\arraystretch}{1.15}
\scriptsize
\begin{tabular}{lc}
\toprule
\textbf{Parameter} & \textbf{Value} \\ 
\midrule
Laser wavelength & 1550 nm \\ 
Pulse width & 50 ns \\ 
Pulse repetition rate & 5 kHz \\ 
Sampling frequency ($f_s$) & 100 MHz \\ 
Spatial resolution ($dx$) & 5 m \\ 
Gauge length ($gl$) & 10 m \\ 
Fiber length ($fl$) & 1.2 km \\ 
Number of channels ($n$) & 240 \\ 
\bottomrule
\end{tabular}
\end{minipage}
\begin{minipage} {0.53\linewidth}
\centering
\caption{Class distribution of the six events in the $\phi$-OTDR laboratory dataset.}

\label{tab:event_numbers}
\renewcommand{\arraystretch}{1.15}
\scriptsize
\begin{tabular}{lcccc}
\toprule
\textbf{Event type} & \textbf{Number} & \textbf{Label} & \textbf{Train} & \textbf{Test} \\ 
\midrule
Background & 3094 & 0 & 2505 & 589 \\ 
Digging & 2512 & 1 & 2018 & 494 \\ 
Knocking & 2530 & 2 & 2025 & 505 \\ 
Watering & 2298 & 3 & 1853 & 445 \\ 
Shaking & 2728 & 4 & 2183 & 545 \\ 
Walking & 2450 & 5 & 1969 & 481 \\ 
\midrule
\textbf{Total} & \textbf{15612} & -- & \textbf{12553} & \textbf{3059} \\
\bottomrule
\end{tabular}
\end{minipage}

\vspace{-0.5cm}

\end{table}
This dataset provides a reliable benchmark for DAS event classification, with each sample represented as a spatio-temporal segment of $10000$ time samples across $12$ adjacent channels. This configuration captures both temporal and spatial dynamics of vibration events, while the controlled experimental setup enables reproducible and consistent comparison of feature extraction methods and learning architectures.

\subsubsection{Real-world scenario dataset}

The second dataset is derived from a real-world DAS deployment introduced by Tomasov \textit{et al.}~\cite{tomasov2025comprehensive}.  
It consists of DAS recordings acquired along an underground optical fiber deployed on a university campus (Fig.~\ref{fig:datasets_overview}-2), enabling the monitoring of diverse human and vehicular activities such as walking, running, longboarding, car traffic, as well as security-related events including fence climbing, manhole opening, and fiber manipulation.

Each raw DAS recording is provided as a two-dimensional matrix of size $T \times C$ (time samples $\times$ channels), together with an event bitmap of size $N_w \times C_b$ indicating the temporal and spatial occurrence of events.  
To ensure compatibility with the laboratory dataset and a unified learning pipeline, the raw data were transformed into a standardized window-based representation.  
First, the bitmap channels were aligned to the DAS data using nearest-neighbor interpolation.  
The recordings were then segmented into $N_w$ temporal windows, and each window was assigned an event label based on the corresponding bitmap entries, with non-event windows labeled as background.
After this transformation, the dataset is reorganized into uniformly structured spatio-temporal windows of size $(2053,1700)$, identical to those of the laboratory dataset, allowing seamless integration into the proposed \textit{DAStatFormer} framework. In total, nine event classes are obtained.  
Table~\ref{tab:real_dataset_dist} reports the resulting class distribution, which remains relatively balanced while reflecting realistic field variability.

\begin{table} [!h]
\centering
\caption{Label distribution in the transformed real-scenario DAS dataset.}
\label{tab:real_dataset_dist}
\renewcommand{\arraystretch}{0.9}
\scriptsize
\begin{tabular}{lcccc}
\toprule
\textbf{Event type} & \textbf{Number of windows} & \textbf{Label} & \textbf{Train}& \textbf{Test}  \\
\midrule
Car & 1094 & 0 & 875 & 219 \\
Construction & 494 & 1 & 395 & 99 \\
Fence interaction & 571 & 2 & 457 & 114 \\
Longboard & 682 & 3 & 546 & 136 \\
Manipulation & 529 & 4 & 423 & 106 \\
Open/Close & 128 & 5 & 102 & 26 \\
Regular activity & 738 & 6 & 590 & 148 \\
Running & 1000 & 7 & 800 & 200 \\
Walking & 1455 & 8 & 1164 & 291 \\
\midrule
\textbf{Total} & \textbf{6691} & -- & \textbf{5352} & \textbf{1339}\\
\bottomrule
\end{tabular}
\end{table}

The final dataset structure thus consists of nine labeled categories, each stored in separate directories, 
allowing for straightforward integration into the preprocessing and feature extraction pipelines of the proposed \textit{DAStatFormer}.

\subsection{Results and Discussion}

This section presents the experimental results obtained by the proposed \textit{DAStatFormer} on both the open $\phi$-OTDR laboratory dataset~\cite{cao2023open} and the real-scenario dataset from Tomasov \textit{et al.}~\cite{tomasov2025comprehensive}.  
The performance is compared with representative approaches from the literature, including classical machine learning, CNN-based, hybrid, and Transformer-based models.  
The model configuration and the training/evaluation protocol are summarized in Tables~\ref{tab:cfg_model} and~\ref{tab:cfg_train}, respectively.

\vspace{-0.5cm}

\begin{table} [!h]
\centering
\captionsetup{font=small}
\begin{minipage} {0.48\linewidth}
\centering
\scriptsize
\caption{DAStatFormer configuration.}
\label{tab:cfg_model}
\renewcommand{\arraystretch}{1.15}
\begin{tabular}{@{}ll@{}}
\toprule
\textbf{Component} & \textbf{Setting} \\
\midrule
Branches & 3 (time,waveform,spectral) \\
Input window & $x \!\in\! \mathbb{R}^{B \times L \times F}$, $F{=}24$ \\
Transformer encoders & $N{=}8$ blocks per path \\
Model width & $d_{\text{model}}{=}256$ \\
FFN hidden dim & $d_{\text{hidden}}{=}128$ \\
Multi-head attention & $h{=}8$, $q{=}8$, $v{=}8$ \\
Dropout & 0.2 (MHA/FFN path) \\
Mask & Enabled (to encoder) \\
Classifier & Linear $\rightarrow$ Softmax \\
\bottomrule
\end{tabular}
\end{minipage}\hfill
\begin{minipage} {0.48\linewidth}
\centering
\scriptsize
\caption{Training protocol settings.}
\label{tab:cfg_train}
\renewcommand{\arraystretch}{1.15}
\begin{tabular}{@{}ll@{}}
\toprule
\textbf{Item} & \textbf{Setting} \\
\midrule
Loss & Cross-Entropy \\
Optimizer / LR & Adam, $\mathrm{LR}{=}1{\times}10^{-4}$ \\
Batch size / Epochs & 32 / 100 \\
Gradient clip & $\lVert g \rVert_2 \le 1.0$ \\
Validation split & 10\% (random) \\
Test interval & Every 5 epochs \\
Seed & 30 \\
Device & CUDA:0 \\
PE / Mask & PE: True; Mask: True \\
\bottomrule
\end{tabular}
\end{minipage}
\end{table}

\begin{figure} [!h]
\centering
\begin{minipage} {0.5\textwidth}
    \centering
    \includegraphics[width=5.2cm]{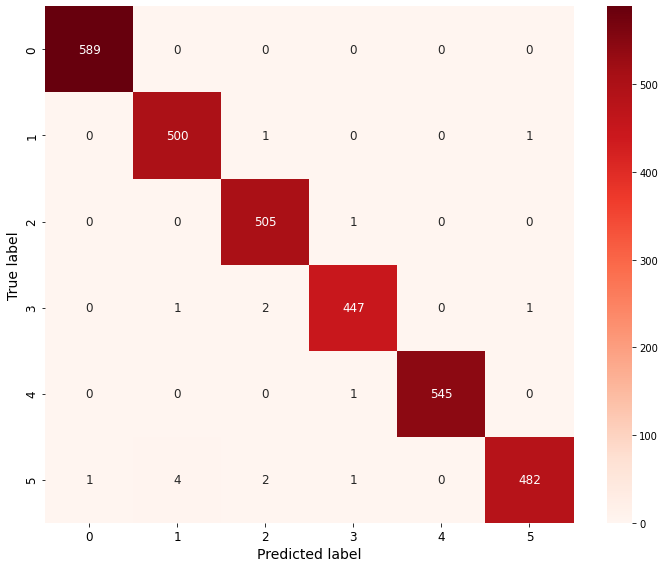}
    \caption*{\scriptsize(a) Laboratory dataset}
\end{minipage}
\hfill
\begin{minipage} {0.42\textwidth}
    \centering
    \includegraphics[width=5.2cm]{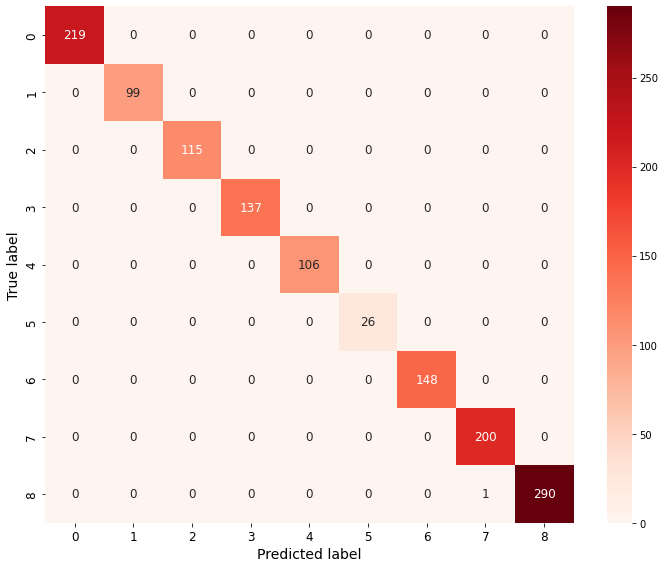}
    \caption*{\scriptsize(b) Real-scenario dataset}
\end{minipage}
\caption{Confusion matrices of \textit{DAStatFormer} on (a) the laboratory dataset and (b) the real-scenario test sets.}

\label{fig:conf_matrices}
\end{figure}

Figure~\ref{fig:conf_matrices} shows the confusion matrices of \textit{DAStatFormer} on both datasets.  
The left matrix corresponds to the laboratory dataset with six event classes (background, digging, knocking, watering, shaking, walking), whereas the right matrix represents the real-scenario dataset containing nine classes (vehicle and human activities recorded on a university campus).  
In both cases, \textit{DAStatFormer} demonstrates strong discriminative ability with very few misclassifications.

\vspace{-0.5cm}

Tables~\ref{tab:comparison_lab} and~\ref{tab:comparison_real} present a comparative evaluation of \textit{DAStatFormer} against existing models on both datasets.  
For clarity, the tables are displayed side-by-side to highlight performance differences between laboratory and field conditions.

\begin{table} [!h]
\centering
\caption{Performance comparison on the open $\phi$-OTDR laboratory dataset.}
\label{tab:comparison_lab}
\renewcommand{\arraystretch}{1.2}
\setlength{\tabcolsep}{7pt}
\scriptsize
\begin{tabular}{lcc}
\toprule
\textbf{Model} & \textbf{Accuracy}$\uparrow$ & \textbf{NAR}$\downarrow$ \\ 
\midrule
SVM~\cite{cao2023open} & 82.6 & 0.184 \\ 
2D CNN~\cite{cao2023open} & 94.0 & 0.0124 \\
CNN–BiLSTM~\cite{wu2020novel} & 97.91 & 0.0042 \\ 
XGBoost~\cite{meng2020research} & 94.26 & 0.0074 \\ 
DeepViT (image-based)~\cite{zhou2021deepvit} & 93.10 & 0.0687 \\ 
DASFormer~\cite{li2025dasformer} & \textbf{99.60} & 0.0000 \\ 

\midrule
\textbf{DAStatFormer (ours)} & 99.48 & \textbf{0.0000} \\
\bottomrule
\end{tabular}
\end{table}
\vspace{-0.4cm}

As shown in Table~\ref{tab:comparison_lab}, \textit{DAStatFormer} achieves an accuracy of \textbf{99.48\%} on the open $\phi$-OTDR laboratory dataset, closely matching the performance of DASFormer~\cite{li2025dasformer} (99.6\%),with a nuisance alarm rate (NAR) of 0.00.  
While the difference in accuracy (0.12\%) is negligible, \textit{DAStatFormer} requires lower computational resources.  
The confusion matrices for the two datasets obtained with \textit{DAStatFormer} are shown in Fig.~\ref{fig:conf_matrices}.  
Unlike DASFormer’s complex dual-tower architecture (over 6~million parameters), \textit{DAStatFormer} leverages compact multidomain statistical embeddings, reducing training time, GPU memory consumption (115 MB allocated), and inference cost by approximately an order of magnitude.  
These results confirm that \textit{DAStatFormer} maintains high detection precision while achieving higher computational efficiency, making it well suited for real-time DAS event recognition.

\vspace{-0.4cm}
\subsubsection{Ablation Study on Feature Domains}  

To assess the contribution of each feature domain, an ablation study was conducted on the $\phi$-OTDR laboratory dataset.  
Four configurations were evaluated: using only time-domain features, only spectral-domain features, only waveform-domain features, and finally the full multidomain combination as employed in the proposed \textit{DAStatFormer} framework.  
The goal of this analysis is to quantify how each domain contributes to the classification performance and to verify the complementarity between them.  

\begin{figure}[!h]
\centering
\includegraphics[width=8cm]{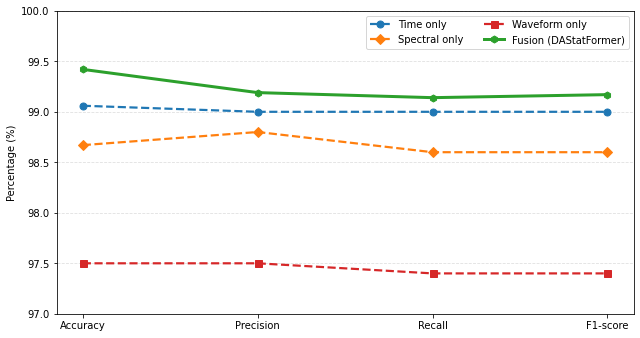}
\caption{ Performance comparison of \textbf{DAStatFormer} across three individual domain features (Time, Spectral, and Waveform) and their fusion.}
\label{fig:ablation}
\end{figure}
\vspace{-0.4cm}

As shown in Fig.~\ref{fig:ablation}, the multidomain fusion significantly improves both accuracy and generalization compared to single-domain configurations. 
While the time-domain features individually achieve high accuracy ($99.1\%$), their combination with spectral and waveform attributes yields the best overall performance ($99.48\%$) with zero false alarms (NAR = $0.00$) and a very low false negative rate (FNR = $0.0004$). 
Interestingly, the multidomain configuration also results in a reduced training time ($84.37$~min), while maintaining a low inference cost ($0.94$~ms/sample). 
These results confirm that the three feature domains provide complementary information: time-domain descriptors capture global signal dynamics, waveform features emphasize impulsive patterns, and spectral features encode frequency-dependent behaviors. 
Their fusion thus enables a more complete and discriminative representation of DAS events.
\begin{table}[t]
\centering
\captionof{table}{Performance comparison and inference time on real-field DAS data.}
\label{tab:comparison_real}
\renewcommand{\arraystretch}{1.3}
\setlength{\tabcolsep}{5pt}
\scriptsize
\begin{tabular}{lcccccc}
\toprule
\textbf{Model} & \textbf{Acc.}$\uparrow$ & \textbf{Prec.}$\uparrow$ & \textbf{Rec.}$\uparrow$ & \textbf{F1}$\uparrow$ & \textbf{Inf. Time}$\downarrow$ & \textbf{NAR} $\downarrow$\\ 
\midrule
CNN (Tomasov et al., 2025)~\cite{tomasov2025comprehensive} & 91.4 & 85.7 & 88.2 &  86.8 & -- & -- \\ 
DASViT1D ~\cite{dione2025intrusion} & 93.5 & 91.93 & 90.80 & 91.29 & 13 & 0.005
 \\ 
\textbf{DAStatFormer (ours)} & \textbf{99.93}& \textbf{99.94}& \textbf{99.97}& \textbf{99.96} & \textbf{4.13}   & \textbf{0.000}\\ 
\bottomrule
\end{tabular}
\end{table}

On the real-field DAS dataset, \textit{DAStatFormer} achieves state-of-the-art performance with 99.93\% accuracy and a macro F1-score of 99.96\%, while producing virtually no false alarms (NAR = 0.00).
This represents a substantial improvement over the CNN baseline of Tomasov \textit{et al.}~\cite{tomasov2025comprehensive} (91.4\% accuracy, 86.8\% F1) and the Transformer-based DASViT1D~\cite{dione2025intrusion} (93.5\% accuracy, 91.29\% F1).
In addition to its superior accuracy, \textit{DAStatFormer} is significantly more efficient, requiring only 4.13 ms per sample for inference, compared to 13 ms for DASViT1D.
These results highlight the effectiveness of multidomain statistical features combined with gated multibranch attention, enabling robust and scalable DAS event recognition under real-world conditions.

\section{Conclusion}
\label{sec:conclusion}

This paper presented \textit{DAStatFormer}, a hybrid multibranch Transformer framework that combines multidomain statistical feature extraction with gated attention mechanisms for scalable and interpretable DAS-based event recognition. By leveraging compact temporal, waveform, and spectral descriptors, the proposed approach effectively captures both local dynamics and long-range dependencies while drastically reducing data dimensionality and computational cost. Extensive experiments conducted on two DAS datasets covering laboratory, and real-scenario conditions demonstrate strong robustness and generalization, with accuracies exceeding 99.4\% and low false-alarm rates. Compared to existing CNN-, hybrid-, and Transformer-based methods, including DASFormer, \textit{DAStatFormer} achieves a more favorable trade-off between accuracy, efficiency, and interpretability. Future work will investigate extensions toward multi-sensor configurations, self-supervised learning strategies, and lightweight implementations for real-time and edge-based DAS monitoring systems. 

\section*{Acknowledgment}
This project was funded by the French State as part of France 2030 operated by ADEME, and funded by the European Union – NextGenerationEU.


\begin{thebibliography}{21}

\bibitem{tomasov2023enhancing}
Tomasov, A., Zaviska, P., Spurny, V., Dejdar, P., Munster, P., Horvath, T., Klicnik, O.:
Enhancing perimeter protection using $\phi$-OTDR and CNN for event classification.
In: \textit{Optical Fiber Sensors}, pp.~W4--39. Optica Publishing Group (2023)

\bibitem{hibert2017automatic}
Hibert, C., Provost, F., Malet, J.-P., Maggi, A., Stumpf, A., Ferrazzini, V.:
Automatic identification of rockfalls and volcano-tectonic earthquakes at the Piton de la Fournaise volcano using a Random Forest algorithm.
\textit{Journal of Volcanology and Geothermal Research} \textbf{340}, 130--142 (2017)

\bibitem{jia2019knn}
Jia, H., Liang, S., Lou, S., Sheng, X.:
A k-Nearest Neighbor Algorithm-Based Near Category Support Vector Machine Method for Event Identification of $\phi$-OTDR.
\textit{IEEE Sensors Journal} \textbf{19}, 3683--3689 (2019)

\bibitem{ghazali2024state}
Ghazali, M.F., Mohamad, H., Nasir, M.Y.M., Hamzh, A., Abdullah:
State-of-the-art application and challenges of optical fibre distributed acoustic sensing in civil engineering.
\textit{Optical Fiber Technology} \textbf{87}, 103911 (2024)

\bibitem{wu2020novel}
Wu, H., Yang, M., Yang, S., Lu, H., Wang, C., Rao, Y.:
A novel DAS signal recognition method based on spatiotemporal information extraction with 1DCNNs-BiLSTM network.
\textit{IEEE Access} \textbf{8}, 119448--119457 (2020)

\bibitem{tejedor2017machine}
Tejedor, J., Macias-Guarasa, J., Martins, H.F., Pastor-Graells, J., Corredera, P., Martin-Lopez, S.:
Machine learning methods for pipeline surveillance systems based on distributed acoustic sensing: A review.
\textit{Applied Sciences} \textbf{7}(8), 841 (2017)

\bibitem{bublin2021event}
Bublin, M.:
Event detection for distributed acoustic sensing: Combining knowledge-based, classical machine learning, and deep learning approaches.
\textit{Sensors} \textbf{21}, 7527 (2021)

\bibitem{cao2023open}
Cao, X., Su, Y., Jin, Z., Yu, K.:
An open dataset of $\phi$-OTDR events with two classification models as baselines.
\textit{Results in Optics} \textbf{10}, 100372 (2023)

\bibitem{vaswani2017attention}
Vaswani, A., Shazeer, N., Parmar, N., Uszkoreit, J., Jones, L., Gomez, A.N., Kaiser, {\L}., Polosukhin, I.:
Attention is all you need.
In: \textit{Advances in Neural Information Processing Systems (NeurIPS)}, vol.~30 (2017)

\bibitem{li2025dasformer}
Li, Y., Qin, Y., Hu, L., Wu, H., Yu, K.:
DASFormer: A long sensing sequence classification and recognition model for phase-sensitive optical time domain reflectometers.
\textit{IEEE Sensors Journal} (2025)

\bibitem{jiang2018event}
Jiang, F., Li, H., Zhang, Z., Zhang, X.:
An event recognition method for fiber distributed acoustic sensing systems based on the combination of MFCC and CNN.
In: \textit{International Conference on Optical Instruments and Technology: Advanced Optical Sensors and Applications},
SPIE \textbf{10618}, 15--21 (2018)

\bibitem{liu2021gated}
Liu, M., Ren, S., Ma, S., Jiao, J., Chen, Y., Wang, Z.:
Gated transformer networks for multivariate time series classification.
\textit{arXiv preprint arXiv:2103.14438} (2021)

\bibitem{meng2020research}
Meng, H., Wang, S., Gao, C., Liu, F.:
Research on recognition method of railway perimeter intrusions based on $\phi$-OTDR optical fiber sensing technology.
\textit{IEEE Sensors Journal} \textbf{21}(8), 9852--9859 (2020)

\bibitem{zhou2021deepvit}
Zhou, D., Kang, B., Jin, X., Yang, L., Lian, X., Hou, Q., Feng, J.:
DeepViT: Towards deeper Vision Transformer.
\textit{arXiv preprint arXiv:2103.11886} (2021)

\bibitem{jestin2020integration}
Jestin, C., Hibert, C., Calbris, G., Lanticq, V.:
Integration of machine learning on distributed acoustic sensing surveys.
\textit{Copernicus Meetings Technical Report} (2020)

\bibitem{xie2023label}
Xie, Y., Wang, M., Zhong, Y., Deng, L., Zhang, J.:
Label-free anomaly detection using distributed optical fiber acoustic sensing.
\textit{Sensors} \textbf{23}, 4094 (2023)

\bibitem{tomasov2025comprehensive}
Tomasov, A., Zaviska, P., Dejdar, P., Klicnik, O., Horvath, T., Munster, P.:
Comprehensive Dataset for Event Classification Using Distributed Acoustic Sensing (DAS) Systems.
\textit{figshare} (2025)

\bibitem{chen2020review}
Chen, X., Liu, W., Zhang, J., Song, Z.:
A review of feature extraction and selection for vibration-based fault diagnosis of rotating machinery.
\textit{Mechanical Systems and Signal Processing} \textbf{160}, 107894 (2021)

\bibitem{dione2025intrusion}
Dione, M., Lonlac, J., Lecoeuche, S., Fleury, A.:
Intrusion Pattern Recognition in DAS Using Multi-Domain Features and a Transformer Network.
In: \textit{Proceedings of the IEEE 15th International Conference on Pattern Recognition Systems} (2025)

\bibitem{song2025facial}
Song, D., Liu, C.:
A facial expression recognition network using hybrid feature extraction.
\textit{PLOS ONE} \textbf{20}(1), e0312359 (2025)

\bibitem{guzhov2021esresnet}
Guzhov, A., Raue, F., Hees, J., Dengel, A.:
ESResNet: Environmental sound classification based on visual domain models.
In: \textit{Proceedings of the 25th International Conference on Pattern Recognition (ICPR)},
pp.~4933--4940 (2021)

\end{thebibliography}
\end{document}